\documentclass[10pt,twocolumn,letterpaper]{article}

\usepackage{iccv}
\usepackage{times}
\usepackage{epsfig}
\usepackage{graphicx}
\usepackage{amsmath}
\usepackage{amssymb}

\usepackage[dvipsnames]{xcolor}
\usepackage[normalem]{ulem}

\usepackage[ruled,noend,linesnumbered]{algorithm2e}

\usepackage[pagebackref=true,breaklinks=true,colorlinks,bookmarks=false]{hyperref}

\newcommand{\Figref}[1]{Fig.~\ref{#1}}
\newcommand{\Tblref}[1]{Table~\ref{#1}}
\newcommand{\Secref}[1]{Sec.~\ref{#1}}
\newcommand{\Algoref}[1]{Algo.~\ref{#1}}

\newcommand{\spacedEight}{\qquad\qquad\quad}
\newcommand{\spacedFive}{\qquad\qquad\qquad\qquad\ \ \ }
\newcommand{\spacedFour}{\qquad\qquad\qquad\qquad\qquad\quad}
\newcommand{\spacedThree}{\qquad\qquad\qquad\qquad\qquad\qquad\qquad\qquad}

\iccvfinalcopy % *** Uncomment this line for the final submission

\def\iccvPaperID{10723} % *** Enter the ICCV Paper ID here

\ificcvfinal\fi

\begin{document}

%%%%%%%%% TITLE
\title{BIDCD - Bosch Industrial Depth Completion Dataset}

% \title{Industrial Object RGBD Dataset for Depth Completion and Synthetic to Depth-Camera Domain Adaptation}

\author{Adam Botach, Yuri Feldman\\
Technion - Israel Institute of Technology\\
Haifa, Israel, 3200000 \\
{\tt\small {[botach, yurif]}@cs.technion.ac.il}
\and
Yakov Miron, Yoel Shapiro, Dotan Di Castro\\
Bosch Center for Artificial Intelligence \\
Haifa, Israel, 3508409\\
{\tt\small {[Yakov.Miron, Shapiro.Yoel, Dotan.DiCastro]}@bosch.com}
}

\maketitle

\begin{abstract}

    We introduce BIDCD - the Bosch Industrial Depth Completion Dataset. BIDCD is a new RGBD dataset of metallic industrial objects, collected with a depth camera mounted on a robotic manipulator. The main purpose of this dataset is to facilitate the training of domain-specific depth completion models, to be used in logistics and manufacturing tasks. 
    We trained a State-of-the-Art depth completion model on this dataset, and report the results, setting an initial benchmark.
    Further, we propose to use this dataset for learning synthetic-to-depth-camera domain adaptation. Modifying synthetic RGBD data to mimic characteristics of real-world depth acquisition could potentially enhance training on synthetic data. 
    For this end, we trained a Generative Adversarial Network (GAN) on a synthetic industrial dataset and our real-world data. Finally, to address geometric distortions in the generated images, we introduce an auxiliary loss that promotes preservation of the original shape.
    The BIDCD data is publicly available at https://zenodo.org/communities/bidcd.

\end{abstract}

\section{Introduction}

\begin{figure*}
    \begin{center}
        \includegraphics[width=0.99\linewidth]{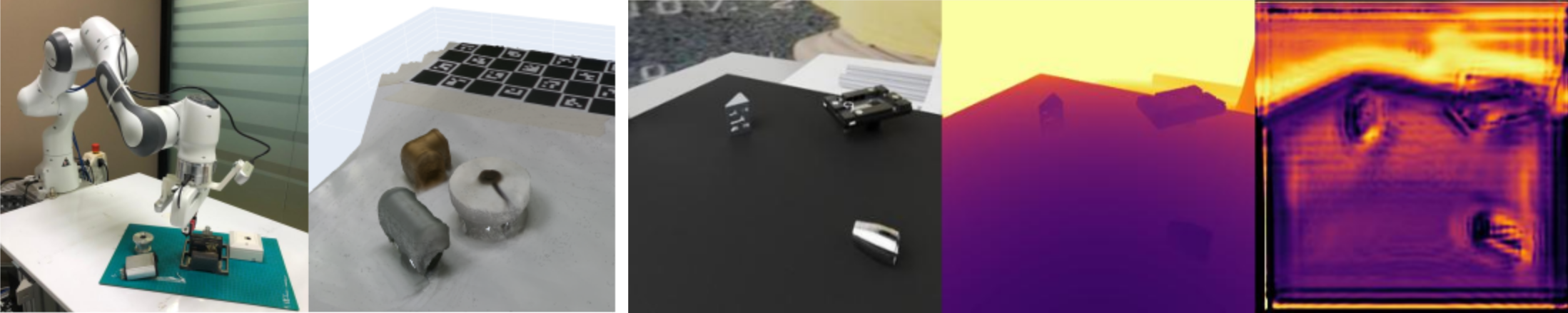}
        \vspace{2mm}
        $(a) \spacedFive (b) \spacedFive (c) \spacedFive (d) \spacedFive (e)$
        \includegraphics[width=0.99\linewidth]{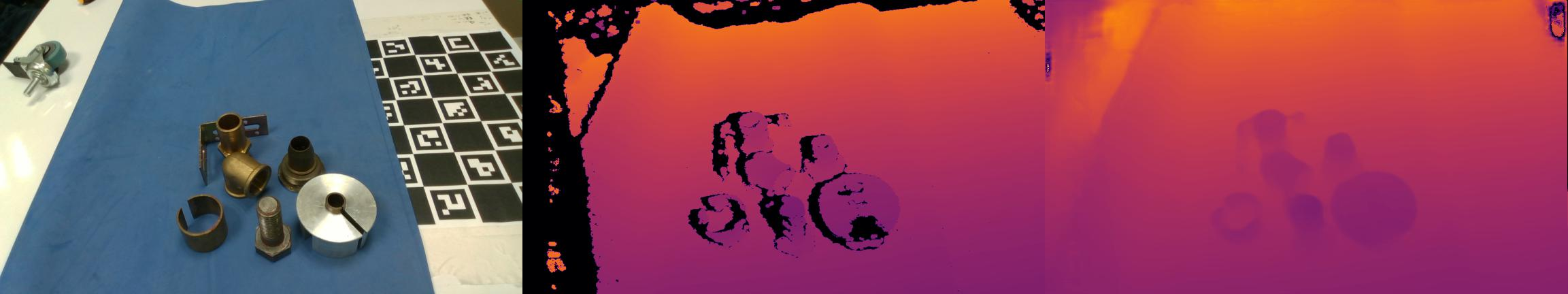}
        $(f) \spacedThree (g) \spacedThree (h)$
    \end{center}
    \caption{
        Our contributions. 
        Top-left: Data Collection. 
	(a) Our data collection setup consists of a Panda arm (Franka-Emika) and two RealSense D415 (Intel) depth cameras, mounted at different angles and distances from the end-effector. 
	(b) A fused 3D mesh, generated from multiple views. Projecting the mesh back to the camera view points yields the ground truth estimation.
        Bottom: Depth Completion. (f) RGB. (g) Raw depth - the peripheral holes are out-of-range regions. 
	(h) Depth completion result - the internal holes have been closed, and the table top has been extrapolated outwards.
        Top-right: Synthetic-to-Depth-Camera Domain Adaptation. (a) Synthetic RGB. (b) Synthetic depth - without any holes or artifacts. 
	(c) Fake depth - the generator adds internal and peripheral holes.
    }
    \label{fig:intro}
\end{figure*}

Robotic manipulation is an important component of the industry 4.0 revolution, both for logistic tasks such as bin-picking and assembly tasks such as peg-in-hole. Operation in unstructured environments requires perception, most commonly implemented with a camera for data acquisition feeding a computer vision model. Perception models need to provide 3D information for task and motion planning, which is typically done in six or more Degree-of-Freedom (DOF). 
Digital RGB cameras have been used to estimate 3D geometry using Structure-from-Motion (SfM) \cite{sfm_2017}, single-frame monocular depth estimation \cite{monocular_Godard_2019_ICCV}, and 3D pose estimation of known objects \cite{tremblay2018corl:dope}. 

In contrast to RGB cameras, distance sensors can provide a direct measurement of 3D geometry. In outdoor applications, the most notable one being self-driving cars, multiple different modalities have been utilized for depth sensing, including LiDAR, Radar, and Ultra Sound (US). 
For indoor applications, a highly prevalent modality for 3D acquisition is depth cameras, which provide RGBD images \cite{rgbd_datasets}. These dense depth maps have been successfully employed for grasp prediction of unknown objects \cite{mahler2019_dexnet4, zeng2019tossingbot}. 
Depth cameras can be implemented in several different ways, including stereo vision, structured light, and Time-of-Flight (ToF), but all these methods suffer from large errors and loss of information ("holes") under optical reflections and deflection. This pitfall is prominent in industrial settings, which routinely include reflective metallic objects. 

% Depth completion
The shortcomings of depth sensors have been previously addressed with depth completion \cite{depth_comp_unit,Ma19icra,NLSPN,plugNplay19,sajjan2019cleargrasp} on single frames. 
Robotic applications often use mounted cameras, which present the opportunity to utilize multiple views and applying volumetric fusion methods, such as the Truncated Signed Distance Function (TSDF) \cite{tsdf_1996, yak}. 
A preliminary requirement for TSDF is the registration between the different points of view.
However, industrial applications tend to be confounded with a multitude of artifacts, which may in turn cause the registration to fail. 
For this reason and additional operational considerations, single-frame solutions seem to be a more viable alternative.
% Perhaps this could be mitigated with additional pre-processing, e.g. with a reflection detector, the likes of \cite{RAU_2018_ECCV}.

Depth completion can be thought of as a type of guided interpolation. The goal is to generate a dense depth map using a partial sampling (raw depth), guided by visual RGB cues such as perspective, occlusions, object boundaries, and surface normals. 
Depth completion datasets include RGB and raw-depth inputs and an estimated dense Ground-Truth (GT) depth map. In driving applications, the depth is typically obtained with LiDAR \cite{kitti}. In contrast, depth cameras are predominant when imaging indoor scenes \cite{Silberman:NYUv2,dai2017scannet}, household objects \cite{ycb}, and other applications \cite{rgbd_datasets}. 
These two acquisition systems pose distinct challenges for depth completion, suggesting that they warrant different solutions.
LiDAR provides sparse depth maps where the values of the samples are typically considered to be sufficiently accurate. Thus, generating a dense depth map is largely a question of partitioning the influence regions around the depth samples using RGB cues such as boundaries.
Depth cameras, on the other hand, may yield large connected regions with missing or invalid values, for example at window panes \cite{Silberman:NYUv2,dai2017scannet}. Moreover, specular reflections and transparent recyclable objects \cite{sajjan2019cleargrasp} may induce large regions of erroneous values that need to be identified and corrected.

\subsection{Industrial Object RGBD Dataset}

Here, we present an industrial-objects dataset \cite{bidcd:online} collected with a depth camera mounted on a robotic manipulator. 
To generate the ground truth, we recorded each scene from approximately 60 Points-of-View (POV) and fused them to produce an estimated GT depth, as described in the methods section. 
Using multiple POVs improve the coverage of the scene, mainly by resolving occlusions and reflections. We configured the cameras to a depth resolution of 0.1 mm and a dynamic range of 0.2-1.2 m, following characteristic range and accuracy requirements for robotic manipulation. 
Consequently, certain peripheral regions of the GT image are left empty (zeros) if they are out-of-range for all POVs. The GT might also have internal holes stemming from localized surface reconstruction failures, typically due to poor observability.
Notably, most RGBD datasets were designed to avoid depth acquisition issues (for example, see \cite{ycb}). In support of industrial applications, we created a dataset rich in highly reflective metallic objects. Our raw depth maps have holes that take on average $~8.5\%$ of the image area, within the pertinent workspace. For details, please refer to \Secref{methods:fusion}.

% \subsection{Depth Completion Model}
We trained a depth completion model on our dataset, to provide a baseline for this task.
Most deep models for depth completion are based on auto-encoders \cite{Hinton504_autoencoder}, which are modified to consume RGBD and output a corrected depth map \cite{depth_comp_unit}. 
% a bit redundant, but allows self-referencing
In \cite{Ma19icra}, the authors added UNet connections \cite{UNet} and achieved State-Of-The-Art (SOTA) performance at that time. In \cite{feldman2020depth}, we took the model from \cite{Ma19icra} and carried out an extensive architecture search in order to enhance it. 

In this work, we trained the NLSPN model from \cite{NLSPN}. The authors of the latter study achieved depth-completion SOTA on the datasets of \cite{kitti} and \cite{Silberman:NYUv2}. 
The NLSPN model uses an auto-encoder to predict several intermediate maps and passes them to a Convolutional Spatial Propagation Network \cite{CSPN, CSPNpp} for iterative refinement of the depth prediction.
Further details can be found in \Secref{methods:dc}. The results are reported in \Secref{results:dc}.

\subsection{Synthetic-to-Real Domain Adaptation}

Synthetic RGBD data and annotations can be generated relatively cheaply and have been successfully used to train deep models for a variety of tasks \cite{semantic_pose_2015,scenenet_rgbd,DeepHPS,unknown_objs2019}.
However, the inherent domain gap between synthetic and real data is very likely to impede such efforts. 
In \cite{abr_2021}, the authors generated photorealistic industrial RGBD data, but their flawless depth maps are radically different from the depth camera outputs we observed in real-world settings.

Our dataset can also be used in the opposite direction, that is, to learn a transformation from GT to depth-camera, simulating the appearance of raw depth. This type of style-transfer could also be applied to synthetic industrial datasets, such as those described in \cite{ABCDataset,abr_2021}.
The authors of \cite{cycgan_depth_da} demonstrated the usefulness of synthetic-to-real-depth domain adaptation for person segmentation.  Here, we trained a newer Generative Adversarial Network (GAN), taken from \cite{cut_2020}, using our raw depth maps as the target domain ("real"). The synthetic data was taken from the dataset presented in \cite{abr_2021}.

Notably, we encountered an issue of shape loss, where objects in the synthetic source image vanished in the fake depth map. It appears that the generator can learn to remove objects and replace them with a hole, or more surprisingly with the depth values of the work surface on which they are laid. 
These depth maps are plausible but contradict the corresponding synthetic RGB image.
The authors of \cite{GANeratedHands_CVPR2018} resolved a similar issue with a geometric consistency loss, which improved the preservation of pertinent features from synthetic RGB images in the fake ("real") images.
We introduce an alternative auxiliary loss for preserving the objects in the fake depth maps. For more details, see \Secref{methods:gan}.

% \ys{remove if we don't run the sparsifier and generate figures}
% \ys{update crossref if we end up putting it in the appendix}
% An alternative method for synthetic-to-real depth conversion was presented in \cite{Ma18icra}, using heuristics for sparsification. 
% We provide qualitative comparisons of our data-driven method and the method from \cite{Ma18icra} in \Secref{results}.

\subsection{Contributions}

We illustrate our contributions in \Figref{fig:intro} and list them here.

\begin{itemize}
    
    \item Data collection - we randomized over 300 scenes of industrial objects and dozens of POVs, collecting a total of $\sim$33k RGBD images using a wrist-mounted (eye-in-hand) depth camera. The challenging nature of this setup yielded many holes in the raw depth images ($\sim8.5\%$ of the image area).
    
    \item Depth ground-truth - the RGBD images were processed with a customized pipeline for filtering, registration, and Point-Cloud (PCD) fusion.
    The fused mesh was back-projected to the original POV to yield an estimated ground-truth depth map.
    
    \item Depth completion - we trained NLSPN \cite{NLSPN} on our dataset, setting an initial benchmark score.
    
    \item Domain adaptation - harnessing our dataset, we trained an existing GAN model \cite{cut_2020} to enhance the photorealism of a synthetic dataset taken from \cite{abr_2021}. 
    Moreover, to address shape loss and the vanishing of objects in the generated depth maps, we introduce an auxiliary loss (see \eqref{eq_pres_loss}).
    
\end{itemize}

\section{Methods} \label{methods}

In this section, we describe how we created the dataset and present the experiments performed for the depth completion task and synthetic-to-real domain adaptation.

\subsection{Data Collection}

\begin{table}
\setlength{\tabcolsep}{11pt}
\renewcommand{\arraystretch}{1.1} % vertical stretch 
\begin{center} 
\begin{tabular}{ |  c |  c  |  c  | }
    \hline
    radius [mm]     & elevation [deg]   & azimuth [deg]     \\ 
    \hline
    349.7 - 416.8   & 44.4 - 78.2     & -151.3 - 153.4    \\
    \hline
\end{tabular}
\caption{
    POV geometric characteristics. The table provides spherical coordinates, two orientation angles, and the distance to the work surface. The coordinate ranges are given with their 10$^{th}$ and 90$^{th}$ percentiles.
}
\label{tbl:povs}
\end{center}
\end{table}

We randomized scenes of industrial objects scattered within a rectangular workspace on top of a table. Each scene consists of up to 10 objects, predominantly metallic items such as screws, cylinders, and heat sinks.
The data was collected with two RealSense D415 depth cameras, mounted on a Panda robotic arm from Franka-Emika. 
The motion path included 30-70 randomized Points-of-View (POV) with the end-effector poses scattered on a hemisphere, while taking into consideration kinematic limitations of the manipulator. The end-effector was kept oriented towards a shared focal point on the table surface. 
In \Tblref{tbl:povs}, we describe the POV camera-axis elevation and azimuth, and the distance of POV from table surface taken along the camera view axis.

Due to the reflective nature of the scenes, the raw depth includes many holes at highlights and object edges. The reflections also cause misleading artifacts with erroneous depth values, as demonstrated in the middle column in \Figref{fig:gt}. Another contributing factor is partial occlusion, i.e. "shadows". On average, the raw depth holes within the workspace take up $\sim8.5\%$ of the image area.

\subsection{Ground Truth Generation}

\begin{figure*}
    \begin{center}
        \includegraphics[width=0.99\linewidth]{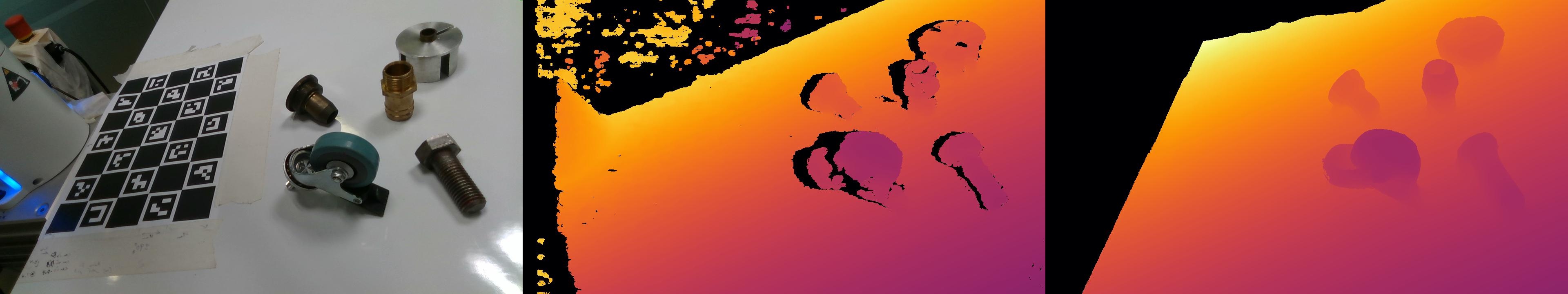}
        \includegraphics[width=0.99\linewidth]{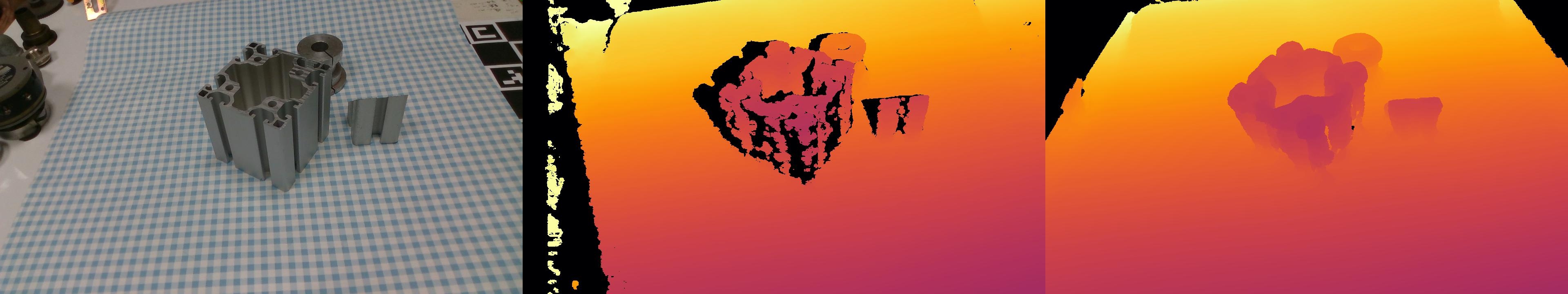}
        \includegraphics[width=0.99\linewidth]{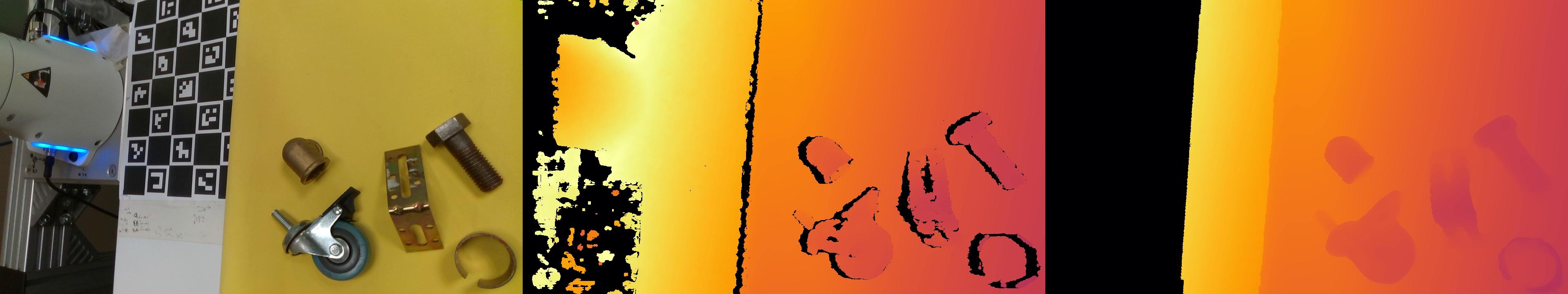}
        \includegraphics[width=0.99\linewidth]{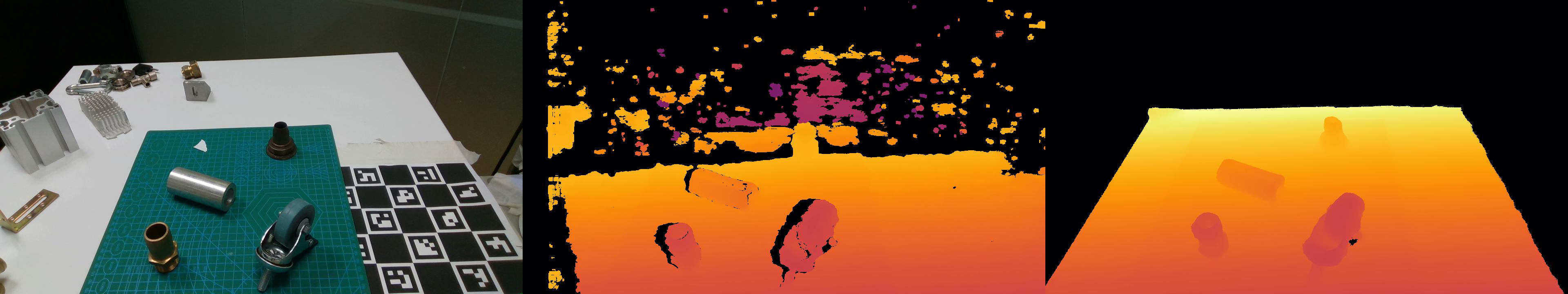}
        $ (a) \spacedThree (b) \spacedThree (c) $
    \end{center}
    \caption{
        Depth Ground Truth. Each row depicts a different sample. From left to right: (a) RGB, (b) raw-depth, (c) GT-depth. In column (b), note how artifacts in the out-of-range regions may appear very close (purple) or very far (yellow). On column (c), we demonstrate how depth-filtering removes artifacts and multi-view fusion adds missing information, and how projecting the pre-defined workspace boundaries onto the depth frames eliminates irrelevant information. On the bottom row column (c), the GT depth has a small inner hole of less than $1\%$ of the image area.
    }
    \label{fig:gt}
\end{figure*}

For each raw depth map, the ground truth is a corresponding dense depth map containing correct distances, approximately within $\pm2mm$ (see \Figref{fig:gt}). Pixels with a distance outside the dynamic range of the camera are designated as invalid (zero values).
To estimate the ground-truth, we applied to each scene the pipeline described below. For clarity, we break it down into three stages, starting from the single POVs raw inputs, followed by multi-view fusion into a 3D mesh, and back to the separate POVs GT.

\subsubsection{Depth Filtering} \label{methods:depth_filt}

Each depth frame was processed individually to remove artifacts, while retaining as much valid information as possible. 
In addition to providing "raw-depth" and "gt-depth" (ground truth), our dataset includes an intermediate result denoted as "cleaned-depth".
The pipeline is depicted in \Algoref{algo:depth} and consists of the following steps: 

\begin{enumerate}

    \item \textbf{Workspace limits} - we project the workspace boundaries onto the raw depth and remove all values outside its perimeter.
    Tall objects positioned near the outer rim may get trimmed because their top protrudes the visual borderline. 
    Hence, we enlarge the actual workspace by a certain margin when applying this step.
    
    \item \textbf{Morphological operations} - looking at the binary mask of the depth values, the objects and tabletop typically form one large blob and false artifacts form separate small blobs. As the main blob may touch artifact blobs, we perform thinning and opening to separate them. We then threshold the blob area to filter out small blobs.
    
    \item \textbf{Outlier removal} - the depth image is first partitioned into a coarse grid and the median depth value is then calculated for each cell. We define a range limit, and depth values exceeding the allowable range from the cell's median are removed. The reasonable range of depth values within a grid cell depends on grid parameters, POV elevation angle, and object height from the table. We used a fixed range threshold, chosen to accommodate the largest possible range within the data.
    
\end{enumerate}

\begin{algorithm} \label{algo:depth}

    \SetAlgoNoLine
    \DontPrintSemicolon
    
    \KwIn{
        $D_{raw} (720 \times 1280), \times$ N;
        $POV \textrm{, homogeneous matrices } H_{i}$;
        $K_{cam} \textrm{, camera intrinsics}$;
        $WSB, \textrm{work space boundaries}$;
        $th_{area}, \textrm{minimal area threshold}$;
        $grid, \textrm{coarse image partitioning}$;
    }
    
    \For{$H_{i} \leftarrow POV$}{
        
        \tcp{Workspace limits}
        $mask_{ws} \leftarrow WSB \textrm{ projection with } H_{i}, K_{cam} $ \\
        initialize $D_{ws}$ \\
        $D_{ws}[mask_{ws}] = D_{raw,i}[mask_{ws}]$ \\
        
        \tcp{Morphological operations}
        $mask_{cln} \leftarrow D_{ws} > 0$ \\
        apply thinning and binary-open on $mask_{cln}$ \\
        apply bwlabel on $mask_{cln} \Rightarrow blobs$ \\
        initialize $D_{cln,i}$ \\
        \For{$blb \leftarrow blobs$}{
            \If{$blb > th_{area}$}{
                $D_{cln,i}[blb] = D_{ws}[blb]$ \\
            }
        }
        
        \tcp{Outlier removal}
        \For{$cell \leftarrow grid$}{
            $m = median(D_{cln}[cell])$ \\
            $mask_{out} = D_{cln,i}[cell] \notin m \pm margin$ \\
            $D_{cln,i}[cell][mask_{out}] = 0$\\
        }
    }
    
    \Return $D_{cln}$, cleaned depth maps of scene
    
    \caption{Depth Filtering}
    
\end{algorithm}

\begin{algorithm} \label{algo:fusion}

    \SetAlgoNoLine
    
    \KwIn{
        $RGB$;
        $D_{cln}$, cleaned depth maps of scene;
        $POV$, homogeneous transform matrices $H_{i}$;
        $th_{dist}$, upper limit on transform distances;
        $th_{clust}$, minimal cluster size threshold
    }
    
    initialize $posegraph$ with N nodes, and no edges \\
    \For{$pov_{i} \leftarrow POV$}{
        $node_{i} \leftarrow pov_{i}$ \\
    }
    
    \tcp{Create edges}
    \For{$i \leftarrow 1 \cdots N$}{
        $RGBD_{cln, i} \leftarrow \left( RGB_{i}, D_{cln,i} \right)$ \\
        \For{$j \leftarrow i + 1 \cdots N$}{
            $RGBD_{cln, j} \leftarrow \left( RGB_{j}, D_{cln,j} \right)$ \\
            $H_{kin} \leftarrow H_{i}^{-1} \cdot H_{j}$ \\
            $H_{icp} \leftarrow ICP \left(RGBD_{cln, i}, RGBD_{cln, j} \right)$ \\
            $H_{delta} \leftarrow H_{kin}^{-1} \cdot H_{icp}$ \\
            $t_{delta} \leftarrow H_{delta} \textrm{ translation vector}$ \\
            \If{$|t_{delta}| < th_{dist}$}{
                $edge_{ij} \leftarrow H_{icp}$
            }
        }
    }
    
    \tcp{Bundle adjustment}
    apply global optimization on $posegraph$ \\
    
    \tcp{TSDF-Volume integration} 
    $mesh \leftarrow TSDF \left( RGBD_{cln} \right)$ \\
    
    $ TriangleClusters \leftarrow \textrm{apply clustering on } mesh $ \\
    \For{$ clst \leftarrow TriangleClusters$}{
        \If{$ clst < th_{clust} $}{
            $ \textrm{remove } clst \textrm{ from } mesh$ \\
        }
    }
    
    \Return $mesh$, 3D model of scene surface
    
    \caption{Multi-View Fusion}
    
\end{algorithm}

\subsubsection{Multi-View Fusion}\label{methods:fusion}

After pre-processing the depth channels, we fused the RGBD frames using the "multiway registration" procedure from \cite{Zhou2018open3d}. For convenience, we outline the steps in \Algoref{algo:fusion}. 
The RGBD frames are organized as a pose-graph, where each pose node corresponds to a POV pose and the robot kinematics are used to initialize it. 
Each edge of the pose-graph, i.e. relative transform between a pair of POVs, is refined with an extension of Iterative-Closest-Point \cite{icp}. 
This stage is prone to errors. While the authors in \cite{Zhou2018open3d} recommend pruning edges with low confidence scores, we found this to be insufficient. We, therefore, replaced replaced the exclusion criteria with a limit on the acceptable ICP update, and prune out transforms deviating beyond a certain distance from the nominal kinematic transforms.
The pose-graph then undergoes global-optimization with the Levenberg-Marquardt algorithm to resolve conflicting relative-transforms (edges). The final pose-graph is used to cast the RGBDs into a shared Point-Cloud (PCD).
The PCD is then triangulated to generate a mesh, using \cite{tsdf_1996}. To remove the remaining artifacts, we applied clustering to the mesh vertices and removed small clusters.

\subsubsection{Back Projection and Hole Filling}\label{methods:back-projection}

To generate GT depth maps, the mesh is projected back to each POV, as described in \Algoref{algo:backproj}. At this stage, the GT depth maps still contain internal holes, which can be filled using RGB-guided interpolation. The authors of \cite{ycb} considered Bi-Lateral Filtering (BLF) \cite{bilat} and the method from \cite{levin2004colorization} and chose to apply the latter. For our dataset, BLF is more suitable, mainly due to the large peripheral out-of-range regions. As shown in \eqref{eq_bilat}, we implement a BLF for RGB distances and pixel-coordinate distances.

\begin{equation} \label{eq_bilat}
\begin{split}
    & \hat D[i, j] =
        \frac{1}{Z[i,j]} \cdot 
        \sum_{uv} D[u,v] \cdot w[u,v] \\
    & Z[i,j] =
        \sum_{uv} w[u,v] \\
    & w[u,v] =
        G_{RGB}[u,v] \cdot G_{Euc}[u,v] \cdot V[u,v] \\
    & G_{RGB}[u,v] =
        \exp \left( -0.5 \cdot \frac
            {||RGB_{ij} - RGB_{uv}||^2}
            { \sigma_{RGB}^2} 
        \right) \\
    & G_{Euc}[u,v] =
        \exp \left( -0.5 \cdot \frac
            {||(i,j) - (u,v)||^2}
            { \sigma_{Euc}^2} 
        \right)
\end{split}
\end{equation}

Where $D$ denotes depth, $V$ is a binary mask of depth validity, the coordinates $uv$ span the filter's support around the central pixel at $ij$, "Euc" is used to denote Euclidean distance in $uv$ coordinates, and $\sigma_{RGB}$ \& $\sigma_{Euc}$ are configuration parameters. 

The filter was applied only to invalid pixels, keeping valid values as-is. To reduce extrapolation into out-of-range regions, we required a minimum occupancy of $50\%$ of valid values within the filter's support.

\begin{algorithm} \label{algo:backproj}

    \SetAlgoNoLine
    
    \KwIn{
        $mesh$; $RGB$;
        $K_{cam} \textrm{, camera intrinsics}$;
        $POV \textrm{, homogeneous matrices } H_{i}$;
        $S \textrm{, support of BLF as square binary kernel}$
    }
    
    \For{$i \leftarrow N$}{
        $D_{GT,i} \leftarrow mesh \textrm{ projection with } H_{i} \textrm{ and } K_{cam}$ \\
        $mask_{v} \leftarrow D_{GT} == 0$ \\
        $mask_{o} \leftarrow mask_{v} * S < 0.5 \cdot \sum S $\\
        $mask_{BLF} = mask_{v} \cdot mask_{o}$ \\
        $D_{BLF} \leftarrow BLF\left(D_{GT,i}, RGB_{i}\right) $ \\
        $D_{GT,i}[mask_{BLF}] \leftarrow D_{BLF}[mask_{BLF}]$
    }
    
    \Return $D_{GT}$, scene ground truth depth maps
    
    \caption{Back Projection and Hole Filling}
    
\end{algorithm}

\subsubsection{Manual Inspection and Retries}

\begin{figure}
    \begin{center}
        \includegraphics[width=1.0\linewidth]{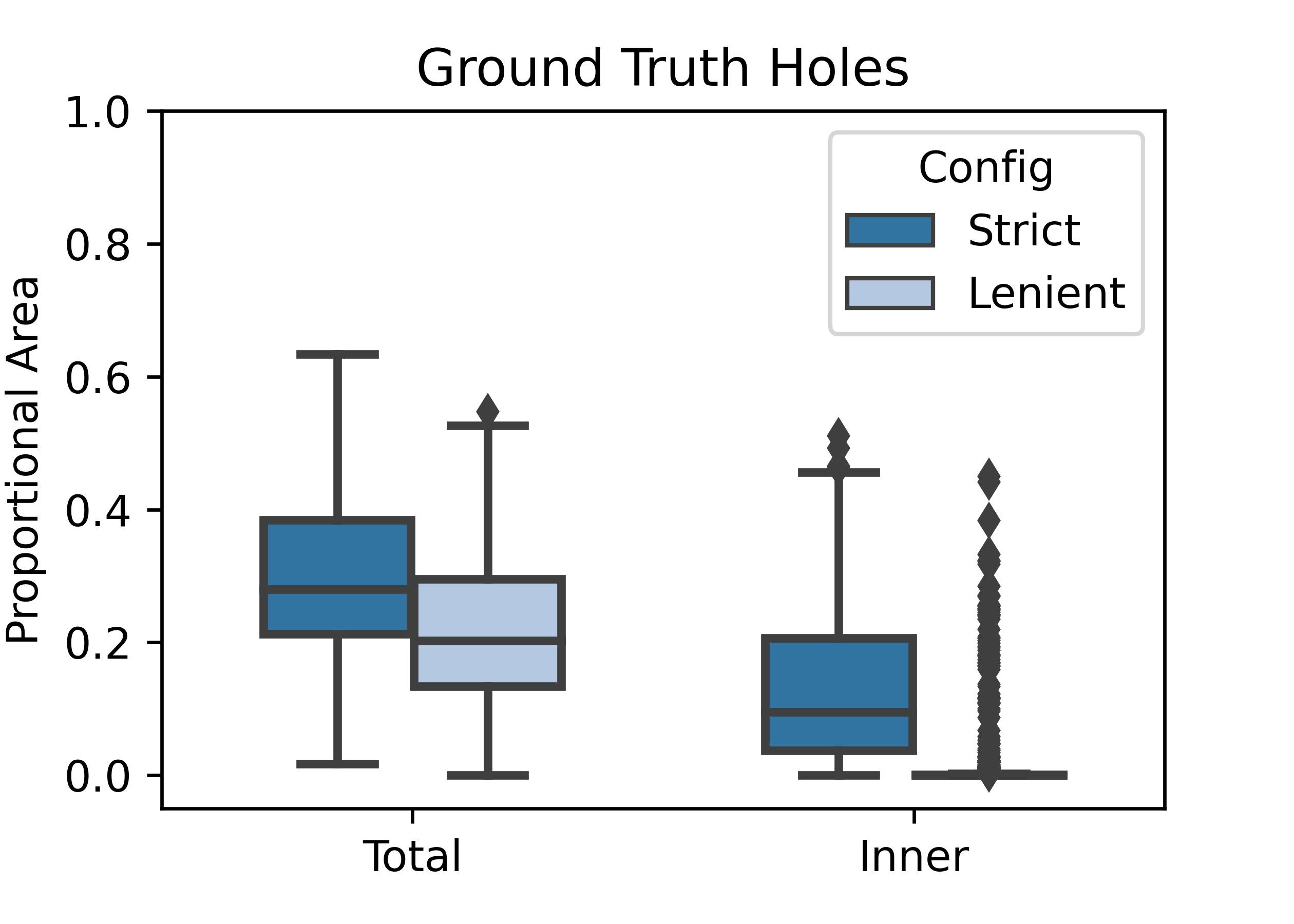}
    \end{center}
    \caption{
        Configuration comparison. We depict the proportional-area of invalid Ground-Truth values, under two different pipeline configurations. 
        The "lenient" configuration retains more information, while the "strict" configuration rejects more artifacts.
        The total size of holes includes peripheral out-of-range regions, which are of less interest.
        Inner holes are only those within the projected workspace (see \Secref{methods:depth_filt}).
    }
    \label{fig:gt_params}
\end{figure}

All the GT depth maps were manually inspected to ensure high fidelity. In seldom cases, we vetted a scene but removed up to $10\%$ of the POV due to persistent artifacts. Scenes that failed curation were reprocessed with stricter parameters, i.e. with more aggressive filtering at the cost of higher information loss. We depict this in \Figref{fig:gt_params}, where we compare the proportion of invalid GT values for 10 random scenes (total of 650 POV) processed with the strict and lenient configurations. For this subset, the raw-depth holes within the projected workspace ("inner") constitute, on average, $8.5\%$ of the image area.

\subsection{Depth Completion} \label{methods:dc}

% The authors released an official implementation on 13 Sep 2020, w MIT license (after Adam)
% https://github.com/zzangjinsun/NLSPN_ECCV20/graphs/contributors

We implemented the model presented in \cite{NLSPN}, consisting of two stages. 
The first stage is an auto-encoder that predicts an initial depth map, affinities, and confidence maps. 
The second stage uses the intermediate outputs to predict the final depth map via iterative non-local spatial propagation. The propagation is akin to convolution with adaptive weights and kernels, determined by confidence and affinities, respectively.
In our implementation, we omitted the confidence prediction to simplify the model. We expect this removal to have a negligible impact since the ablation study in \cite{NLSPN} showed that it reduced errors only by $\sim0.5\%$.
The resolution of the input RGBD frames is 720x1280 pixels, with a 9:16 aspect ratio. The NLSPN model was trained on images reduced to $60\%$ in each dimension, i.e. 432x768 pixels.
% Our implementation is available at \footnote{To be released after review.}.

\subsection{Synthetic-to-Real Depth GAN} \label{methods:gan}

We applied domain adaptation to synthetic data from \cite{abr_2021}, to appear more like our depth maps.
We trained the GAN model from \cite{cut_2020}, an improvement of the model in \cite{CycleGAN2017}.
The generator inputs are synthetic RGBD and it provides the discriminator with a modified depth map. 
The synthetic and real datasets are unpaired, i.e. the fake and real images fed to the discriminator depict different scenes, taken from different POVs.
We convert the synthetic RGB to grayscale, and feed it to the discriminator without modification in conjunction with the fake depth map.
We confirmed with an ablation trial (not shown here) that the additional grayscale information improves geometric consistency. 
The conversion to grayscale reduces the appearance gap between the synthetic and real datasets. Otherwise the discriminator might latch on to extraneous features, such as the work surface color.

\begin{figure*}[!ht]
    \begin{center}
        \includegraphics[width=0.49\linewidth]{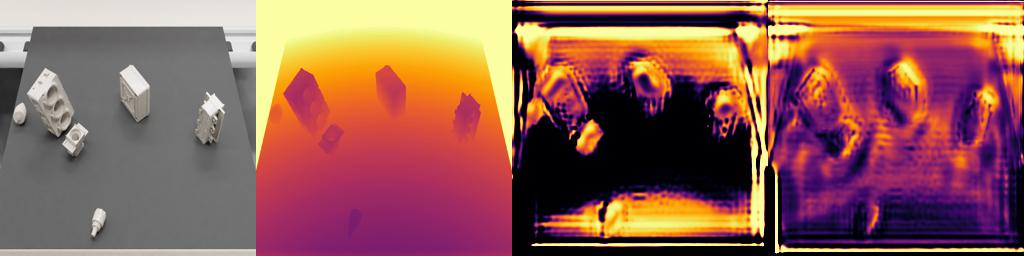}
        % \hspace{0.5mm}
        \includegraphics[width=0.49\linewidth]{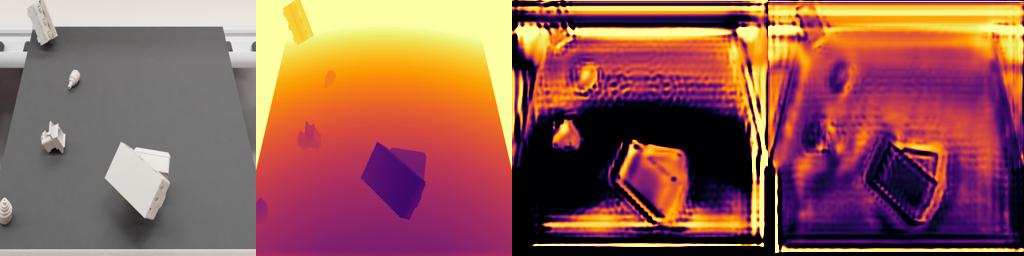}
        \\
        \vspace{0.5mm}
        \includegraphics[width=0.49\linewidth]{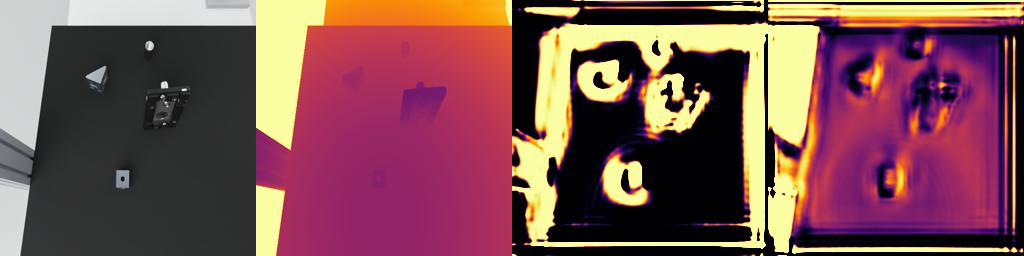}
        % \hspace{1mm}
        \includegraphics[width=0.49\linewidth]{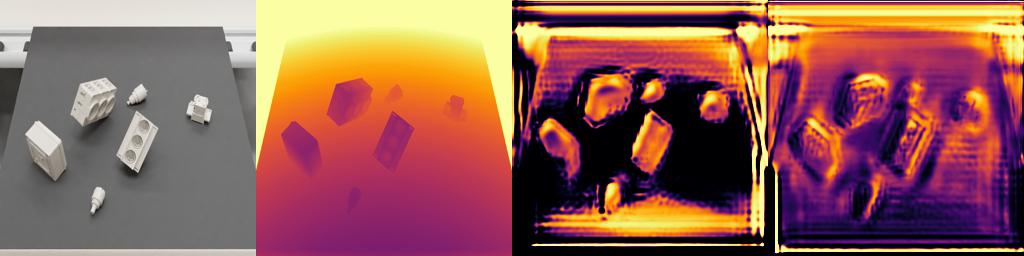}
        $(a) \spacedEight (b) \spacedEight (c) \spacedEight (d) \spacedEight (a) \spacedEight (b) \spacedEight (c) \spacedEight (d)$
    \end{center}
    \caption{
        Synthetic-to-Real Depth Domain Adaptation. We display two samples on each row. Left to right: (a) synthetic RGB, (b) synthetic depth, (c) fake depth w/o preservation loss, (d) fake depth w/ preservation loss $L_{p}$. 
        The generator successfully adds peripheral and internal holes. As expected, the results w/ $L_{p}$ on column (d) appear more realistic.
      }
    \label{fig:GAN_figure}
\end{figure*}

We faced a challenge of vanishing objects in the fake depth (see \Figref{fig:GAN_figure}). 
The generator learned to "remove" objects and replace them with a hole, or the underlying surface that they were placed on. 
We mitigated this by adding a marginal preservation loss $L_{p}$, as shown in \eqref{eq_pres_loss}. We use $D$ to denote depth, and $syn$ for synthetic.
The loss penalizes the generator for modifying the input depth unless the modification is larger than the margin threshold. 
The rationale behind the preservation loss is that holes and artifacts in raw depth images are usually substantially different from their immediate surroundings. The undesired object removal modifies the synthetic depth by object height, which is approximately two orders of magnitude smaller than the desired changes.

\begin{equation}
\begin{split}
    \Delta & = \left| D_{syn} - D_{fake} \right| \\
    margin & = 20^{th} \ percentile \ of \ \Delta \\
    L_{p}[i,j] & = 
        \begin{cases}
            \Delta[i,j] & \Delta[i,j] < margin \\
            0 & otherwise
        \end{cases} \\
\end{split}
\label{eq_pres_loss}
\end{equation}

\begin{figure*}
    \begin{center}
        \includegraphics[width=0.99\linewidth]{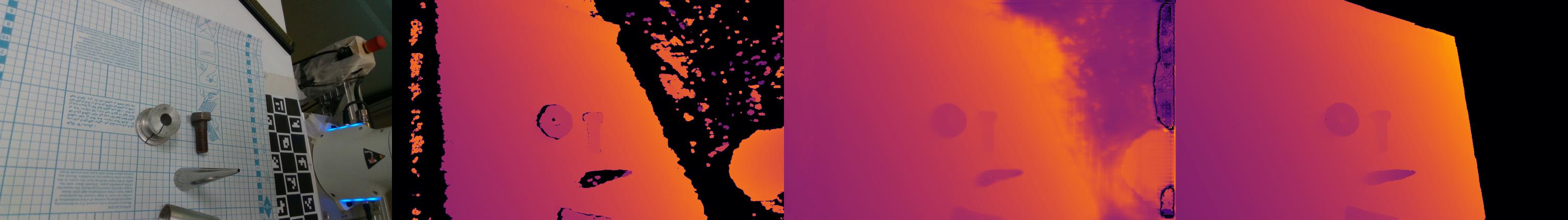}
        \includegraphics[width=0.99\linewidth]{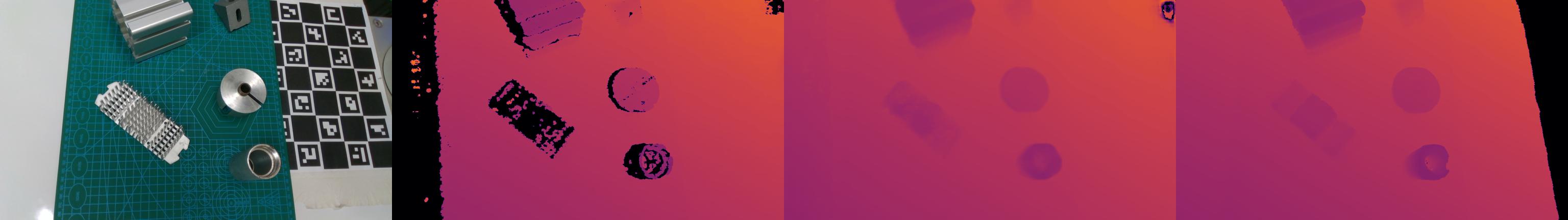}
        $(a) \spacedFour (b) \spacedFour (c) \spacedFour (d) $
    \end{center}
    \caption{
        Depth completion with NLSPN. Each row depicts a different sample. Left-to-right: (a) RGB, (b) Raw-Depth, (c) Depth Prediction, and (d) Ground-Truth. Notice how the model is able to fill shadows and holes on objects. On the peripheral out-of-range regions the artifacts are subdues, and the table top is correctly extended. 
    }
    \label{fig:nlspn}
\end{figure*}

\section{Results} \label{results}

\subsection{Industrial RGBD Dataset}

We were able to collect and generate GT for 319 different scenes of industrial objects. Each scene was acquired from $63.6 \pm 8.1$ POVs (mean $\pm$ 1 std), resulting in over 33k RGBD frames.
Our dataset is organized by scenes. For each POV, we provide the RGB and raw-depth inputs and the corresponding GT-depth. As shown in \Figref{fig:gt}, the GT depth maps have empty peripheral regions which take up $\sim20\%$ of the image area and might include small internal holes taking up less than $1\%$.

We also provide the intermediate results: a mask of the projected workspace and the cleaned-depth (see \Algoref{algo:depth}). To support 3D approaches, we publish a parallel 3D dataset \footnote{anonymized for review} containing camera intrinsic parameters, POVs given as homogeneous matrix, and the colored meshes calculated in \Algoref{algo:fusion}.

% \ys{mention the old single-item dc? or just provide a link in final DS?}
% % details of BIDCD-SI
We applied the ground-truth generation pipeline to previously collected data of single items, which we provide separately at \cite{bidcd:online}, as BIDCD-SI. This dataset consists of single-object scenes, collected with a rig of four static depth cameras. It consists of slightly over 400 scenes and approximately 4000 samples.

\subsection{Depth Completion} \label{results:dc}

\begin{table} 
\renewcommand{\arraystretch}{1.1} % vertical stretch 
\begin{center} 
\begin{tabular}{ | l | r | r |} 
    \hline
                & RMSE      & MAE   \\ 
    \hline
    Raw         & 122.1     & 49.0  \\
    \hline
    NLSPN       & 25.3      & 8.9   \\
    \hline
\end{tabular}
\caption{
    Depth completion errors. The errors are measured in millimeters, given as RMSE and MAE. Raw refers to acquisition errors and NLSPN to depth-completion.
}
\label{tbl:nlspn_errors}
\end{center}
\end{table}

We summarize the test results of the NLSPN model in \Tblref{tbl:nlspn_errors} and show examples of predictions in \Figref{fig:nlspn}. 
The errors are calculated against the GT and only where the GT is valid (non-zero). We report Root-Mean-Square-Error (RMSE) and Mean-Absolute-Error (MAE); the latter is less sensitive to extreme values.
The NLSPN model can reduce the acquisition errors approximately by a factor of five. 
Further work will be necessary to achieve the practical requirements of $\sim$1 mm accuracy.

% NLSPN on our data
% Try adding supervised surface normal estimation

% \ys{testing on NYU is NOT standardized (sparsification) - consider KITTI instead}

\subsection{Synthetic-to-Real Depth Domain Adaptation}

% \begin{figure*}[t]
% \begin{center}
% \includegraphics[width=0.99\linewidth]{images/GAN_Images/043_gt_depth.jpg}
% \end{center}
%   \caption{
%         Domain Adaptation Identity Preservation. Left to right: RGB captured, GT depth, raw depth, fake raw depth w/o preservation loss, and fake raw depth w preservation loss. 
%         We applied the generator to GT depth maps and compared the fake outputs to the corresponding Raw depth. Our preservation loss (extreme right) successfully increases the qualitative similarity to the raw depth (middle column).
%     }
% \label{fig:GAN_figure_gt_to_raw}
% \end{figure*}

We trained a GAN to generate fake depth maps, similar to our real-world depth maps, conditioned on synthetic RGBD images. 
We modified \cite{cut_2020} to consume RGBD rather than RGB images and to generate depth rather than RBG. In addition, we modified the generator to pass a grayscale image of the synthetic image as-is, without modifying it. The synthetic RGBD images were taken from \cite{abr_2021}. The discriminator was tasked with classifying the depth maps as "real" and the generated depth maps as "fake". 

In \Figref{fig:GAN_figure}, we present results of the generative model for four samples. 
It depicts the RGBD inputs from \cite{abr_2021} and two generated fake depth maps, w/ and w/o the preservation loss $L_{p}$ (see \eqref{eq_pres_loss}). 
The predictions on column (c) (w/o $L_{p}$) displays several issues, for example large implausible holes at the table top.
% displays the vanishing objects phenomenon discussed in \Secref{methods:gan}. 
The predictions w/ the preservation loss $L_{p}$ are shown on column (d). The auxillary loss managed to improve geometric consistency with the synthetic RGB and depth.

For more examples, see \Figref{fig:GAN_Supp} in the appendix.

To evaluate this model quantitatively, we input GT images from our dataset as "synthetic" to the generator and compare the $Raw_{fake}$ to the $Raw_{real}$, i.e. the corresponding raw depth. The fake depth is not expected to be identical to the original raw depth, but we do expect some similarity (e.g., size and positioning of holes). We take the absolute difference as error, $err = |Raw_{real} - Raw_{fake}|$, and use its median value $e_{med}$ to estimate the merit of our preservation loss $L_{p}$. Applying $L_{p}$ reduces $e_{med}$ from 46 to 31, a considerable decrease of $\sim30\%$.

% \ys{include comparison to sparsifier ?}

\section{Conclusions} \label{conclusion}

To our knowledge our dataset is the first public RGBD dataset for industrial objects.
To ensure high variability, we used a few dozens of objects with approximately 10 different backgrounds and randomized Points-of-Views.

We trained the NLSPN depth-completion model on our data and report its errors. 
From our experience with grasping models, we asses that the achieved accuracy would be adequate for the more robust models such as \cite{zeng2019tossingbot}, but might be insufficient for more sensitive models such as \cite{mahler2019_dexnet4}.

We performed domain adaptation, synthetic-to-depth-camera, as a means to ameliorate training on synthetic RGBD data. We demonstrate this with a GAN from \cite{cut_2020} and introduce an auxiliary preservation-loss to improve its performance. 

We hope that our contribution will help push forwards robotic manipulation in industrial applications.

%\clearpage
{\small
\bibliographystyle{ieee_fullname}
\bibliography{bib_iccv21}
}

% \newpage
% \clearpage
\appendix
\onecolumn
% supposed to come a s separate file, xxxx-supp.pdf/zip
% might be easier to format as separate file, if we don't use cross-refs

\makeatletter
\setcounter{figure}{0}
\renewcommand{\thefigure}{S\@arabic\c@figure}
\makeatother

\begin{center}
\section{Supplementary Figures}
\end{center}

\newcommand{\wgt}{0.99}

\begin{figure*}[h]
\begin{center}
    \includegraphics[width=\wgt\linewidth]{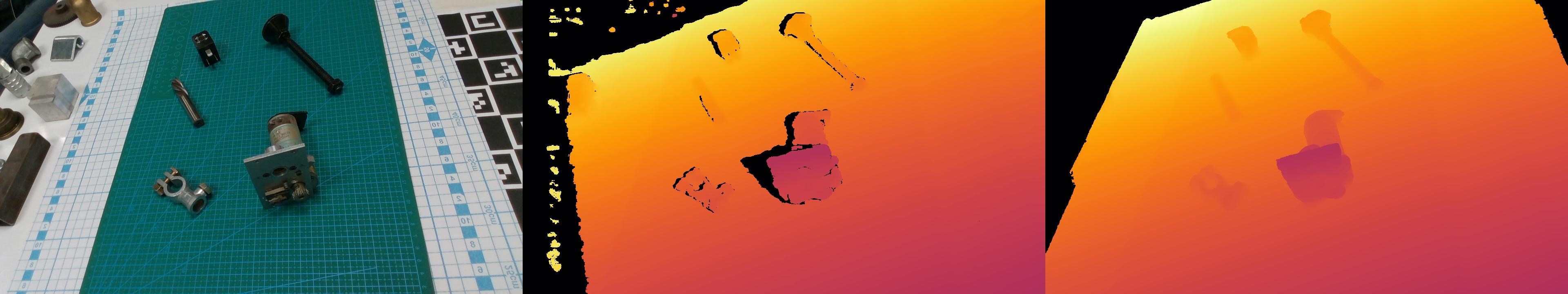}
    \includegraphics[width=\wgt\linewidth]{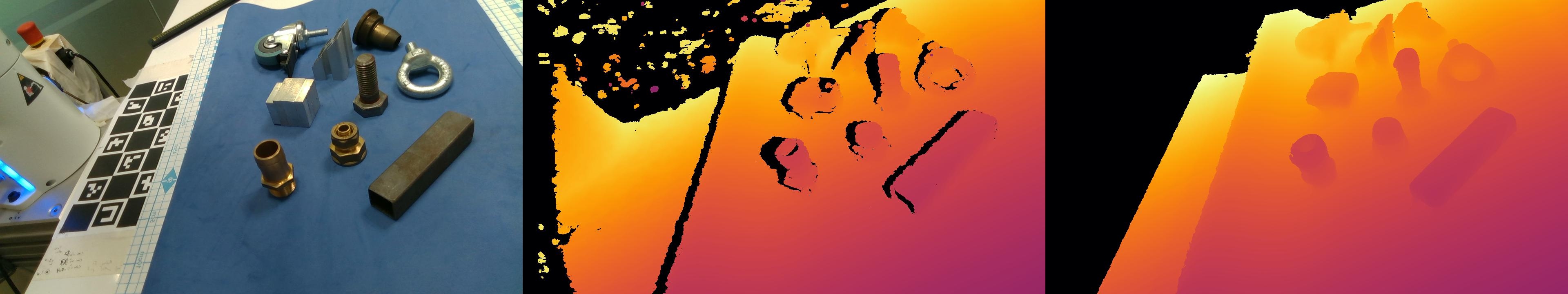}
    \includegraphics[width=\wgt\linewidth]{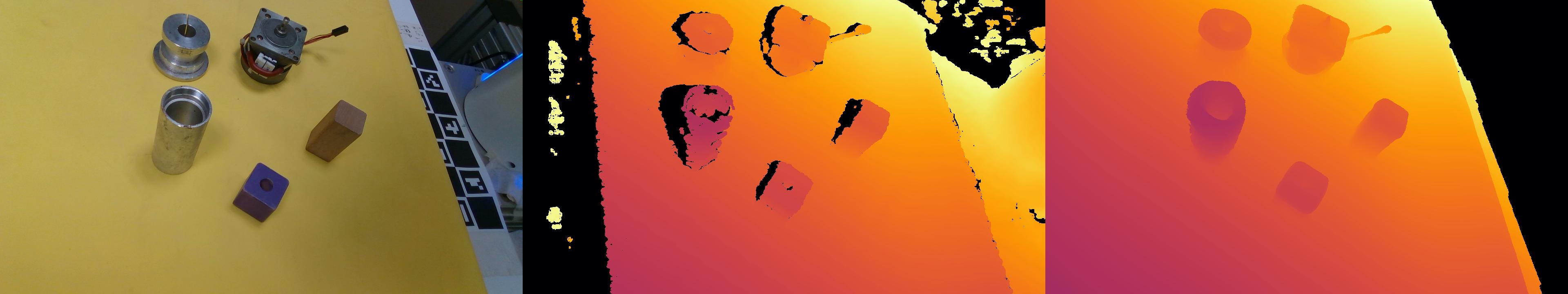}
    \includegraphics[width=\wgt\linewidth]{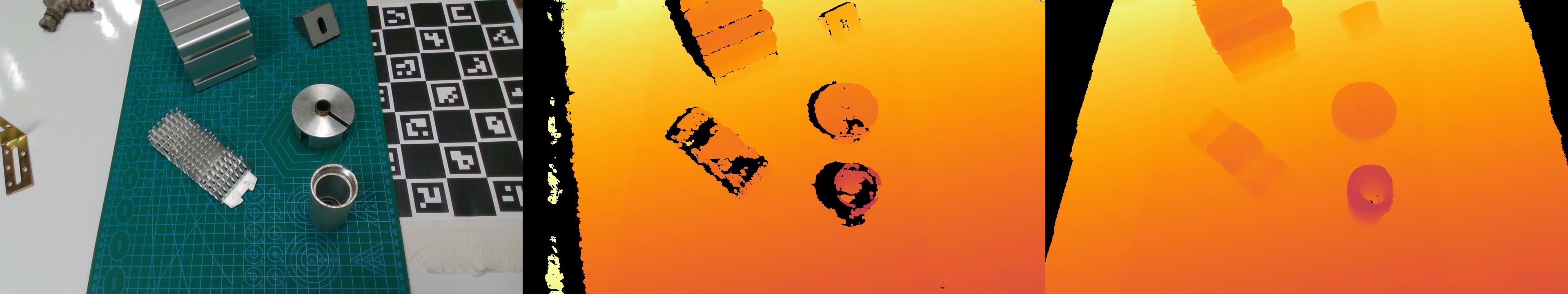}
    \includegraphics[width=\wgt\linewidth]{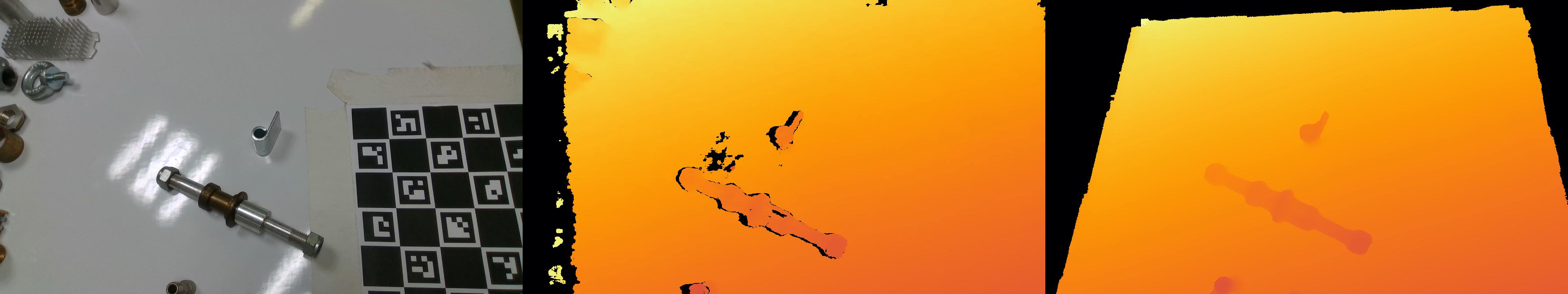}
    \includegraphics[width=\wgt\linewidth]{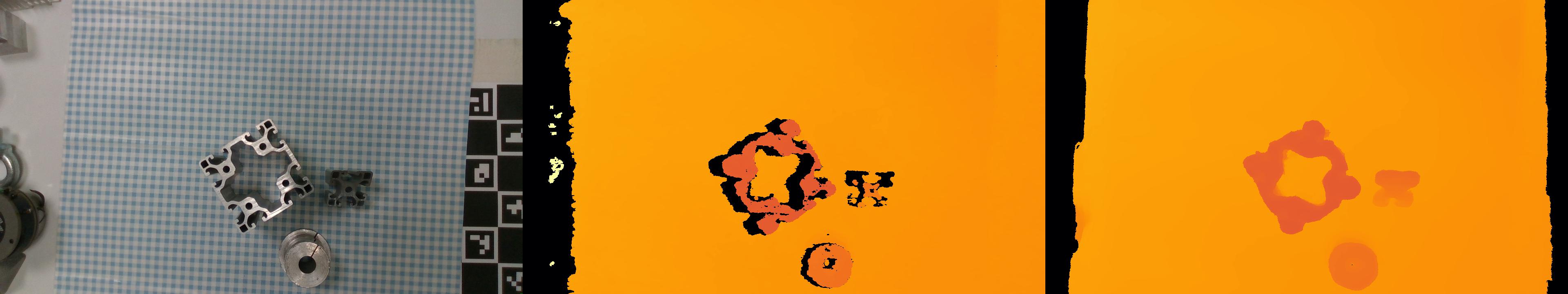}
\end{center}
  \caption{Depth Ground Truth. Each row (left-to-right) depicts an example of RGB, raw-depth, and GT-depth. The raw depth of intricate parts, such as the heat sink and the aluminium profiles is fragmented and missing. The GT pipeline can retrieve most of the information, with some loss of fine detail.}
\label{fig:gt_supp}
\end{figure*}

\begin{figure*}
\begin{center}
    \includegraphics[width=0.99\linewidth]{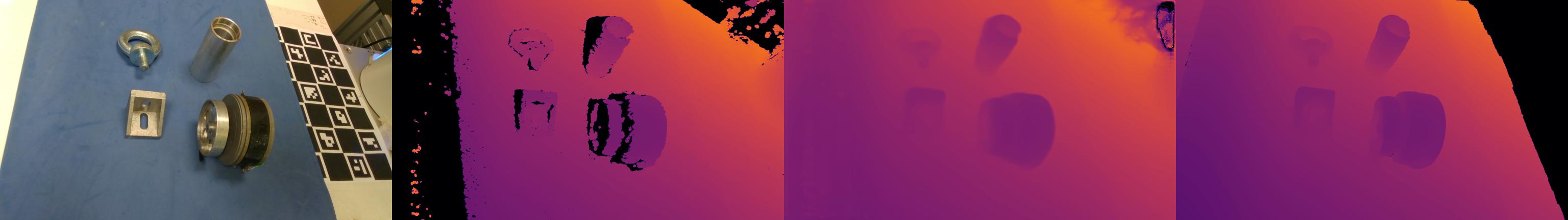}
    \includegraphics[width=0.99\linewidth]{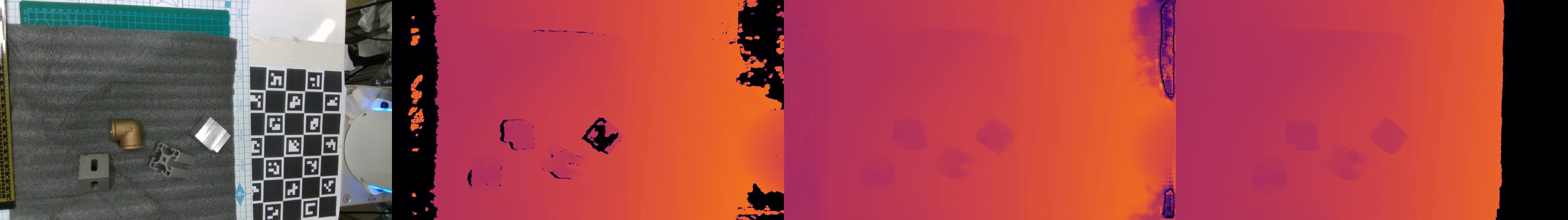}
    \includegraphics[width=0.99\linewidth]{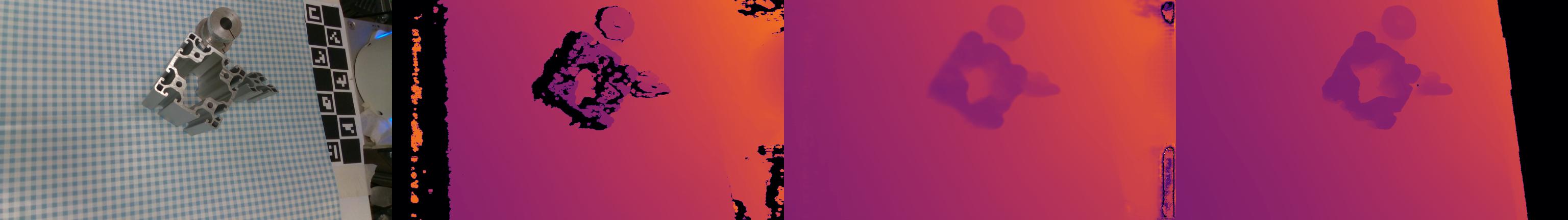}
    \includegraphics[width=0.99\linewidth]{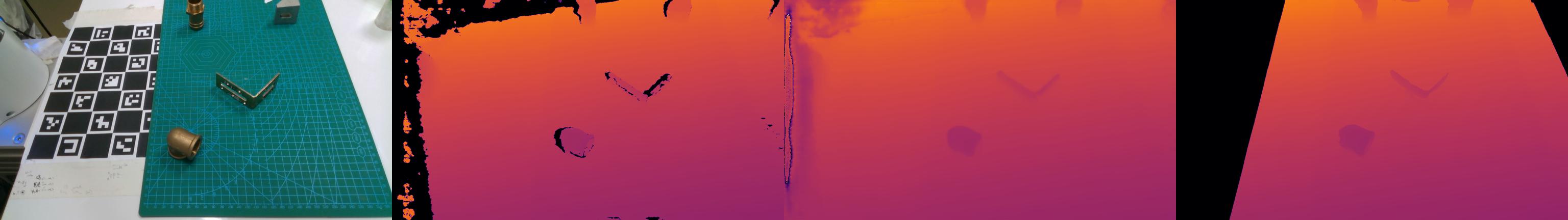}
    \includegraphics[width=0.99\linewidth]{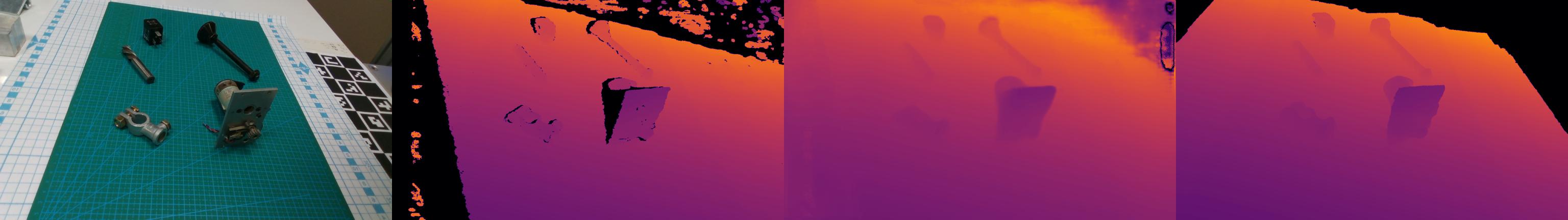}
    \includegraphics[width=0.99\linewidth]{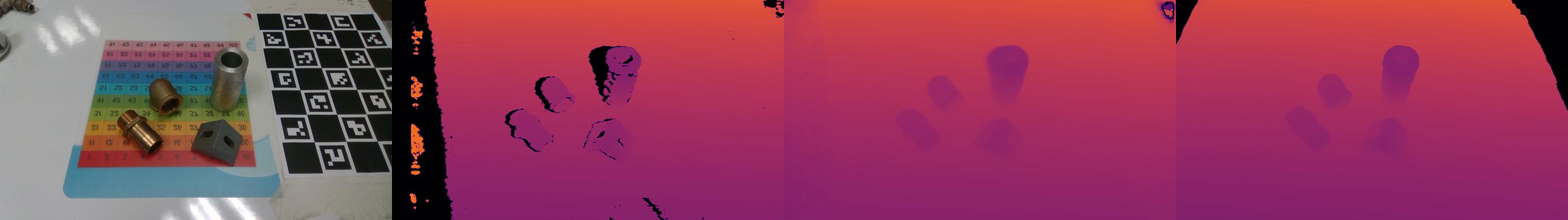}
    \includegraphics[width=0.99\linewidth]{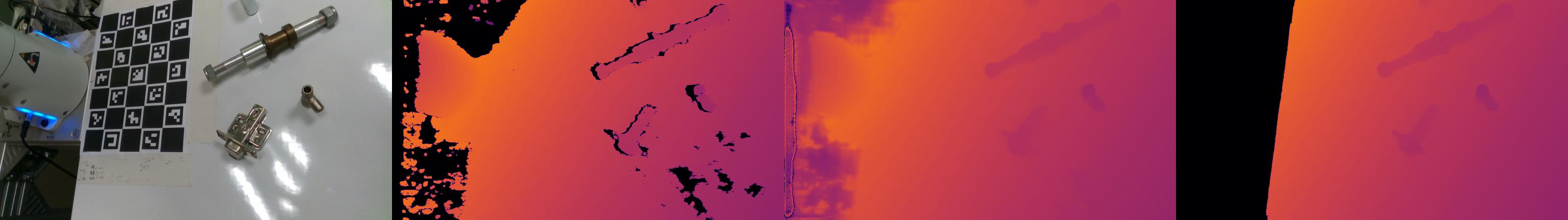}
    \includegraphics[width=0.99\linewidth]{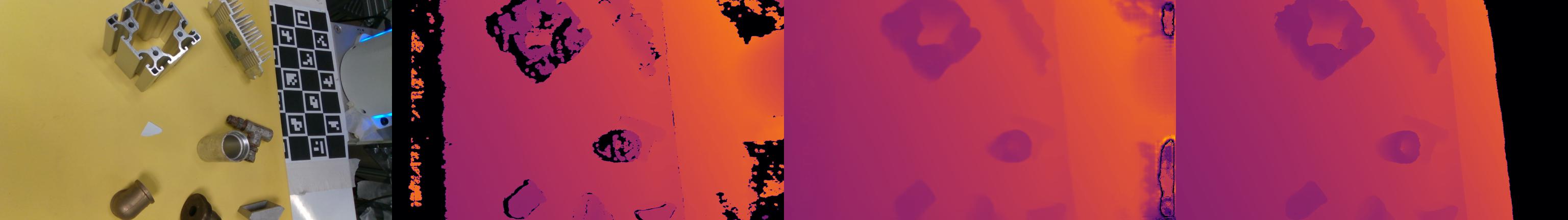}
\end{center}
  \caption{Depth completion with NLSPN. Left-to-right: RGB, Raw-Depth, Prediction, and Ground-Truth. Each row is a different sample. Notice how the model is can fill shadows, holes on objects, and highlights on the work surface. It also partially compensates for the depth-camera dynamic range limits and extrapolates the work surface.}
\label{fig:nlspn_supp}
\end{figure*}

\newcommand{\wgan}{0.65}

\begin{figure*}
\begin{center}
    \includegraphics[width=\wgan\linewidth]{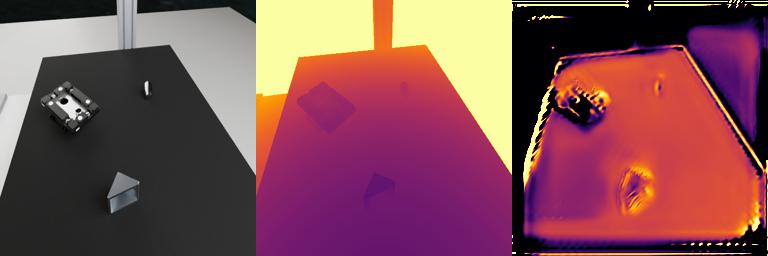}
    \includegraphics[width=\wgan\linewidth]{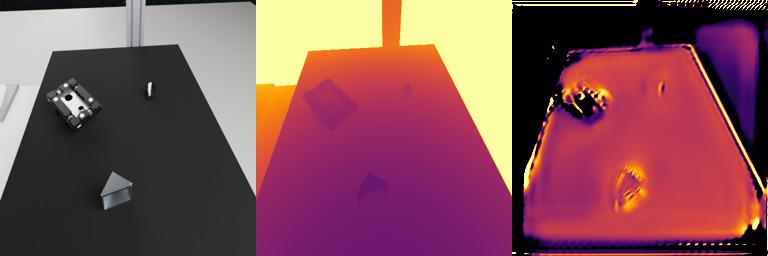}
    \includegraphics[width=\wgan\linewidth]{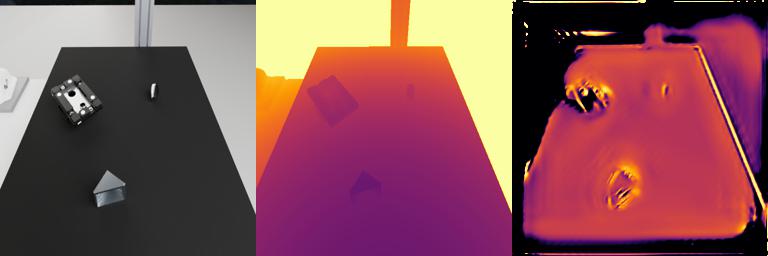}
    \includegraphics[width=\wgan\linewidth]{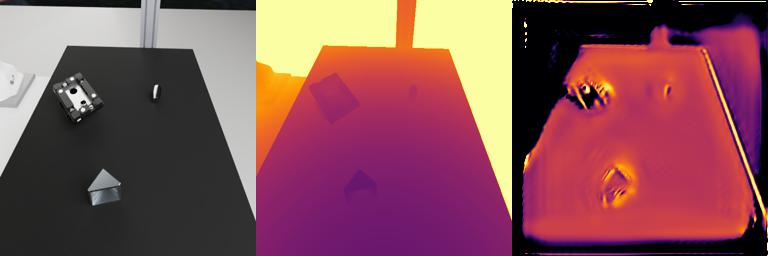}
    \includegraphics[width=\wgan\linewidth]{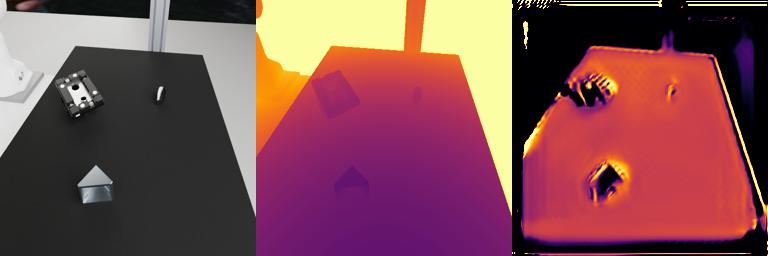}
\end{center}
  \caption{
    Synthetic-to-Real Depth Domain Adaptation. Top-down: consecutive frames,emphasizing consistency. Left to right: synthetic RGB, synthetic depth, fake depth w/ preservation loss.
  }
\label{fig:GAN_Supp}
\end{figure*}

\end{document}